\documentclass[letterpaper]{article} 
\usepackage{aaai24} 
\usepackage{times} 
\usepackage{helvet} 
\usepackage{courier} 
\usepackage[hyphens]{url} 
\usepackage{graphicx} 
\usepackage{natbib} 
\usepackage{caption} 
\usepackage{amsmath,amssymb,amsfonts}
\usepackage{booktabs}
\usepackage{array}
\usepackage{xspace}
\providecommand{\href}[2]{#2}
\newcommand{\projecturl}{%
  \leavevmode
  \pdfstartlink attr{/Border [0 0 0]} user{%
    /Subtype /Link /A << /S /URI
    /URI (https://github.com/pd2714/RL_badminton) >>}%
  {\fontencoding{T1}\fontfamily{ptm}\selectfont
    https://github.com/pd2714/RL\char95{}badminton}%
  \pdfendlink}

\nocopyright

\newcommand{\rev}[1]{#1}
\newif\ifwithappendix
\ifdefined\submissiononly
    \withappendixfalse
\else
    \withappendixtrue
\fi

\title{ShuttleArena: Interpretable Self-Play in Physics-Based Badminton}

\author{
    Peize Ding
}
\affiliations{
    Columbia University\\
    New York, NY 10027, USA\\
    pd2714@columbia.edu
}

\begin{document}

\maketitle

\begin{abstract}
Badminton is a compact but challenging domain for game AI: a player must choose a physically feasible shuttle trajectory, anticipate the opponent's interception, and recover to a court position whose value depends on the opponent's next response.
The central challenge is that shot selection and recovery are not separable: the best recovery depends on the shot-induced opponent response, while the value of the shot depends on whether the hitter can cover the reply.
This paper presents ShuttleArena, a physics-based singles badminton self-play environment that couples continuous shuttle flight, player interception, structured shot generation, and post-shot recovery.
\rev{The policy uses role-conditioned outputs: a masked interception choice on receiver turns and a factorized hitter action over shot azimuth, shot elevation, shot speed, and recovery target, enabling interpretable tactical probes.}
Episodes are single rallies rather than full scored games, and training uses Proximal Policy Optimization (PPO) self-play against a staged checkpoint opponent pool with sparse terminal rally-outcome rewards and a factor-specific recovery update.
Evaluation with frozen checkpoint play, controlled tactical probes, recovery ablations, qualitative rollouts, and a human-data sanity check shows competitive improvement together with interpretable opponent-conditioned changes in shot geometry and recovery behavior.
\rev{The learned policies produce recognizable badminton-like structure while also reflecting the abstractions of the simulator, and the recovery intervention shows that learned recovery behavior is competitively important.}
These results suggest that physics-based racket sports are a useful testbed for interactive digital entertainment AI because they require agents to coordinate execution, positioning, and opponent-relative tactical value.
\end{abstract}

\section{Introduction}

Sports games are attractive domains for game AI.
They combine continuous control, long-horizon consequences, opponent modeling, and visually interpretable behavior.
Badminton is especially revealing: the agent's immediate choice is a shuttle trajectory, but the tactical value of that trajectory depends on whether the opponent can intercept it, what return it induces, and whether the hitter recovers to a useful next position.
\rev{For readers unfamiliar with the sport, singles badminton is played by two opponents who alternate striking a shuttlecock over a net, with at most one stroke per side. A rally ends when the shuttle lands in bounds, is hit out, fails to cross the net, or a player otherwise commits a fault; the rally winner scores a point.}
A smash, clear, lift, drive, or net shot is therefore not good in isolation.
Its value is coupled to the opponent's location, velocity, reaction time, and the hitter's post-shot recovery.

Self-play is a natural training paradigm for such opponent-relative decision making.
It has a long history in games, from TD-Gammon \cite{tesauro1995tdgammon} through AlphaGo and AlphaZero-style self-play \cite{silver2016mastering,silver2018general}.
Modern competitive systems commonly combine policy-gradient optimization such as PPO \cite{schulman2017ppo}, competitive or population-based training \cite{bansal2018emergent,jaderberg2019human}, retrospective skill ratings \cite{elo1978rating,herbrich2007trueskill}, and game-theoretic population views such as policy-space response oracles \cite{lanctot2017unified}.
However, racket-sport agents raise a distinct evaluation problem: aggregate handcrafted metrics, such as average shot speed, may saturate early or fluctuate, while policy quality remains fundamentally relative to the opponent population.

This paper studies badminton self-play as a coupled shot-and-recovery learning problem.
The evaluation focuses not only on final win rate, but also on how the learned policy changes under fixed tactical conditions.
Agents are trained in ShuttleArena with a structured action representation, then evaluated through competitive, controlled, human-data sanity, and ablation probes over frozen checkpoints.
Because the simulator uses badminton-inspired movement, interception, and high-drag shuttle-flight parameters, the emergent shot-and-recovery patterns can be inspected as recognizable badminton tactics without treating the environment as a fully calibrated physical replica.
The primary game-AI contribution is ShuttleArena: an environment that makes the shot--recovery--opponent-response coupling learnable and interpretable.
A structured policy exposes the coupled action factors, and the implementation uses a lightweight recovery-head credit adjustment to support training.
The central empirical claim is that competitive self-play improvement and tactical interpretability should be measured together: the win-rate matrix and rating curve show whether policies improve, controlled shot and recovery probes show what changed behaviorally, a human-data sanity check anchors the patterns against observed rallies, and recovery ablations test whether those behavioral changes matter competitively.

The paper makes six contributions:
\begin{itemize}
    \item ShuttleArena, a physics-based badminton rally environment that couples shot execution, shuttle drag, receiver interception, terminal rally outcomes, and post-shot recovery.
    \item \rev{A role-conditioned structured policy with a 20-way masked interception head for receiver turns and a factorized hitter branch over shot azimuth, shot elevation, shot speed, and recovery target, reducing the $11{,}000$-way hitter choice to $49$ component logits and exposing the recovery decision for controlled analysis.}
    \item A self-play training configuration using PPO, sparse terminal rally-outcome rewards, and a pure-recency stage followed by a broader linear-recency continuation.
    \item An evaluation suite combining frozen round-robin win rates, Bradley--Terry/Elo-style ratings, controlled contact probes, recovery evolution probes, human-data sanity comparison, recovery ablations, and qualitative rally rollouts.
    \item Evidence, across the reported evaluations, of competitive improvement with non-monotonic, matchup-dependent dynamics, together with interpretable changes in shot and recovery behavior.
    \item Human-comparison patterns that are plausible but not identical to observed rallies, and ablation evidence for learned recovery choices and the recovery-head update used in training.
\end{itemize}

Together, these contributions establish ShuttleArena as a compact game-AI testbed in which strategic learning can be inspected directly.
ShuttleArena makes learned behavior visible through shot trajectories, recoveries, and rally outcomes; the structured policy exposes the agent's tactical choices; and the evaluation suite links competitive improvement to opponent-relative changes in play.

\section{Related Work}

\paragraph{Reinforcement learning in games.}
Games have provided central benchmarks for reinforcement learning, including Atari through the Arcade Learning Environment \cite{bellemare2013ale,mnih2015human}, OpenAI Gym \cite{brockman2016openai}, MuJoCo control \cite{todorov2012mujoco}, Obstacle Tower \cite{juliani2019obstacletower}, and multi-agent libraries such as PettingZoo and OpenSpiel \cite{terry2021pettingzoo,lanctot2019openspiel}.
Self-play has produced strong agents in backgammon, Go, chess, shogi, Dota 2, StarCraft II, and multiplayer capture-the-flag settings \cite{tesauro1995tdgammon,silver2016mastering,silver2018general,berner2019dota,vinyals2019alphastar,jaderberg2019human}.
These systems demonstrate the power of opponent-relative learning, but their internal decision variables are often less directly interpretable than the spatial shot and recovery decisions in racket sports.

\paragraph{Policy-gradient and actor-critic methods.}
The implementation builds on policy-gradient and actor-critic ideas \cite{sutton2000policy,konda2000actor}, PPO \cite{schulman2017ppo}, trust-region and generalized advantage estimation (GAE) methods \cite{schulman2015trpo,schulman2015gae}, and reliable open-source RL implementations \cite{raffin2021stablebaselines3}.
Because the policy has explicit shot and recovery factors, the PPO update can assign the recovery factor a separate advantage signal while retaining the same actor-critic backbone.
The counterfactual part of this signal is related in spirit to counterfactual baselines for credit assignment, particularly Counterfactual Multi-Agent (COMA) policy gradients \cite{foerster2018coma}, but applies the comparison to one action factor rather than to one agent in a cooperative multi-agent team.

\paragraph{Structured and hybrid action spaces.}
Badminton actions are naturally structured: a high-level intent determines a trajectory family, while continuous or discretized parameters determine direction, height, speed, and recovery location.
This resembles work on parameterized, branched, and hybrid action spaces \cite{masson2016parameterized,tavakoli2018actionbranching,fan2019hybrid}.
The ShuttleArena policy uses this structure not only for scalability but also for analysis: individual action factors can be probed under fixed tactical states.

\paragraph{Sports and badminton AI.}
Sports environments have become important RL benchmarks, including Google Research Football \cite{kurach2020grf} and multi-agent football self-play studies \cite{jaderberg2017population}.
Badminton-specific AI work has focused on stroke forecasting, tactical datasets, stroke influence analysis, shuttle trajectory reconstruction, and data-driven environments \cite{wang2022shuttlenet,wang2023shuttleset,wang2023shuttleset22,wang2022shotinfluence,liu2022monotrack,ibh2024rallytempose,wang2024coachai,li2026shuttleenv}.
In particular, ShuttleSet provides a human-annotated stroke-level singles dataset for tactical analysis, while ShuttleSet22 frames stroke-level badminton data as a benchmark for stroke forecasting \cite{wang2023shuttleset,wang2023shuttleset22}.
These works provide valuable data-driven models and benchmarks.
\rev{The CoachAI Badminton Environment, CoachAI+/RallyNetV2, and ShuttleEnv are especially complementary to ShuttleArena \cite{wang2024coachai,peng2025coachaiplus,li2026shuttleenv}. CoachAI+/RallyNetV2 operates at the stroke-event level, predicting discrete shot types, landing locations, and player movement from match data, while ShuttleArena executes explicit three-dimensional shuttle trajectories and trains policies through online self-play in a physics-based environment. A head-to-head comparison with a released event-level policy would require a state/action adapter and an inverse-trajectory controller, so its outcome would conflate policy quality with the cross-environment mapping. We therefore do not claim that ShuttleArena outperforms CoachAI+/RallyNetV2 or is superior for human-behavior imitation. The systems address complementary goals: ShuttleArena provides a physics-based platform in which shot execution and recovery can be separately probed and intervened upon, whereas the data-driven systems emphasize human-match sequence modeling and strategy evaluation from recorded rallies.}
The shuttlecock itself is a high-drag projectile whose trajectory differs substantially from parabolic ball flight, motivating explicit drag modeling \cite{cohen2015physicsbadminton,collet2026shuttlecock,nokihara2023future}.

\section{ShuttleArena Environment}

\subsection{State and Court Geometry}

\begin{figure*}[t]
    \centering
    \includegraphics[width=\textwidth]{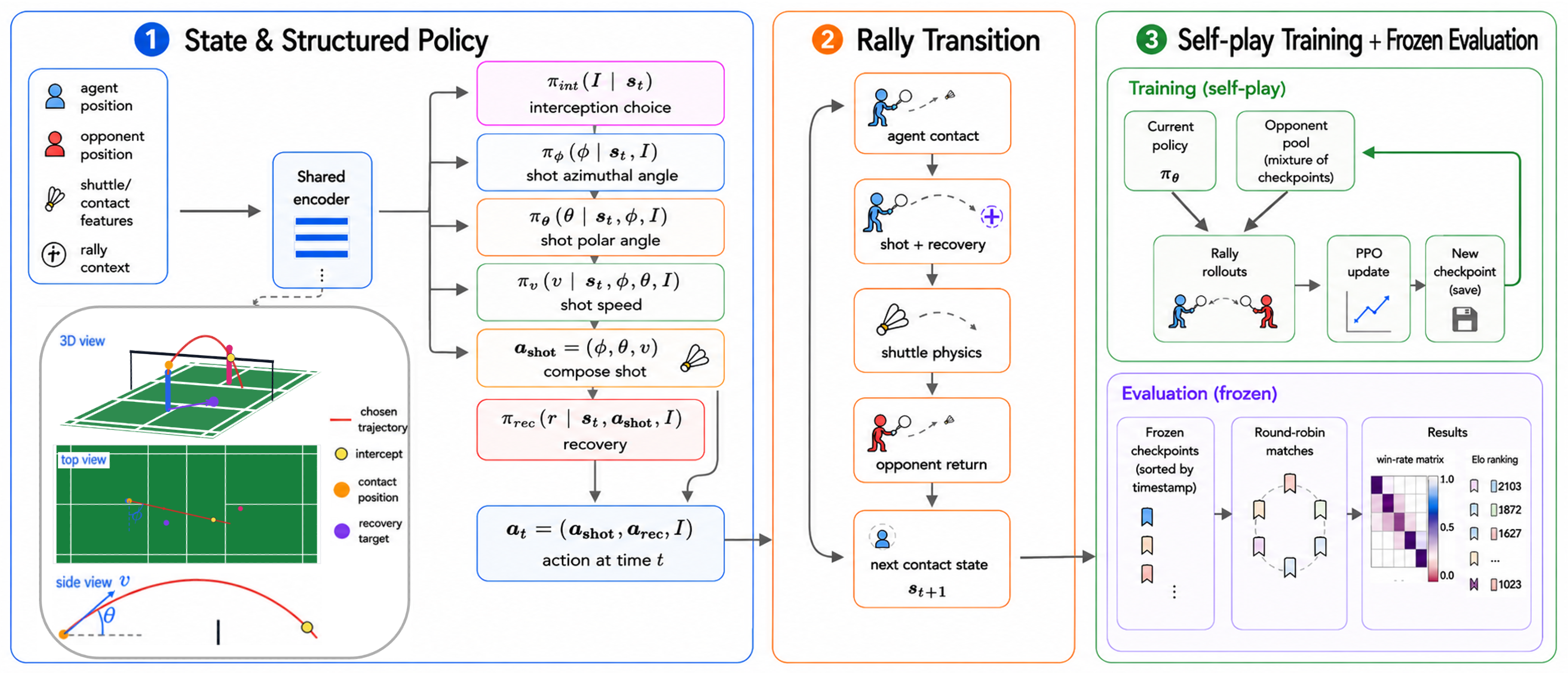}
    \caption{Overview of ShuttleArena, including the state, role-conditioned policy, rally transition, and evaluation pipeline. The policy observes player positions, shuttle/contact features, feasible interception information, and rally context. On receiver turns it selects a feasible interception candidate; after contact, on hitter turns it factorizes the outgoing action into shot azimuth, elevation, speed, and recovery. The left panel compactly shows these consecutive decisions in one receive-and-return chain; they are not sampled jointly in one simulator step. ShuttleArena advances the rally through shuttle physics and opponent response, while PPO self-play trains against a checkpoint mixture and frozen checkpoints are evaluated by round-robin win-rate and rating analyses.}
    \label{fig:overview}
\end{figure*}

As summarized in Figure~\ref{fig:overview}, ShuttleArena represents a singles badminton rally on a two-dimensional court with three-dimensional shuttle flight.
At each policy decision, the current agent observes player positions, shuttle/contact features, feasible interception information when receiving, and rally context.
The policy acts in one of two roles at each simulator step.
As receiver, it selects a feasible interception point for the incoming shot; following successful contact, it acts as hitter and selects the outgoing shot and recovery target.
The formulation is rally-level: one episode is one rally, the terminal reward is the winner of that rally, and the experiments do not simulate full $21$-point games, match scoring, side changes, or service-state effects beyond the randomized rally initialization.
The main simulator, action-space, training, and evaluation settings for the analyzed run are summarized in Table~\ref{tab:runconfig}.
The state includes the hitter and receiver positions, shuttle contact height and velocity-related features, side/server information, and features describing reaction risk and feasible future interception.
For the run analyzed in this paper, the agent trained on the left side, the initial server was randomized, and service $x$ position was randomized.

\subsection{Shuttle Physics, Shot Parameterization, and Interception}
ShuttleArena models shuttle flight as a high-drag projectile under gravity and drag-square dynamics.
For the run analyzed here, the simulator uses separate horizontal and vertical drag coefficients, $k_h=0.20$ and $k_v=0.16$, with numerical integration step $0.01$ s and maximum initial shuttle speed $100.0$ m/s.
The parameters are motivated by badminton physics and produce rapidly decelerating, non-parabolic trajectories that are physically plausible for a high-drag shuttle and more badminton-like than simple ballistic motion.
Appendix~D compares simulated trajectories against BWF (Badminton World Federation) and other reference data across four shot families.

The policy controls this physics model through a velocity-oriented shot representation.
A shot is parameterized by azimuth $\phi$, elevation or polar angle $\theta$, and initial speed $v$:
\begin{equation}
    a^{\mathrm{shot}} = (\phi,\theta,v).
\end{equation}
The trained run discretizes these factors into $11$ azimuth bins, $8$ elevation bins, and $5$ speed bins.
ShuttleArena clips or projects invalid choices to feasible shot outcomes according to the simulator's action-validity logic.
The receiver attempts to intercept the shuttle subject to movement constraints, reaction time, acceleration, racket length, and maximum hitting height.
For learned agents, the receiver chooses among the feasible members of up to 20 candidate interception points sampled along the incoming flight; movement toward the selected point is executed by a prescribed acceleration-limited controller.
The analyzed run uses an accelerated movement model with player speed $5.0$ m/s, acceleration $8.0$ m/s$^2$, racket length $1.6$ m, maximum hitting height $2.6$ m, and reaction time $0.15$ s.
The racket reach is height-dependent, so its horizontal component contracts as contact height approaches the maximum hitting height.
To avoid unrealistically perfect interceptions on shots that leave very little time to react, the simulator also includes a fast-reaction miss model: contacts with less than $0.1$ s of flight time are missed with probability $0.8$; between $0.1$ and $0.5$ s, this probability decreases linearly to zero, and it remains zero thereafter.
A rally terminates when the shuttle is out, cannot be legally intercepted, hits the ground, or reaches the configured maximum rally length.

\subsection{Recovery}
After choosing a shot, the hitter also chooses a recovery target $r$ on a $5 \times 5$ court grid.
As in badminton, this choice is made immediately after hitting the shuttle: the player must decide where to recover while the opponent prepares a return.
ShuttleArena moves the hitter toward this target while the shuttle travels, making recovery important for reaching the next shot and keeping the rally going.
The best recovery is therefore not simply the geometric center of the court, but depends on the outgoing shot and on the opponent response that the shot induces.
ShuttleArena keeps recovery targets a margin away from the sidelines, net, and back boundary.

\section{Structured Policy and Self-Play Training}

\subsection{Policy Factorization}
A monolithic discrete hitter action over all shot and recovery combinations would be large and hard to interpret.
\rev{With the discretization used here, the joint hitter space contains $11\times8\times5\times25=11{,}000$ combinations, whereas the hitter branch predicts $11+8+5+25=49$ component logits. A separate receiver head predicts 20 interception logits, masked to the feasible candidate set.}
The hitter action distribution is factorized according to the physical sequence of the decision:
\begin{align}
    \pi_{\mathrm{hit}}(a_t \mid s_t)
    &=
    \pi_{\phi}(\phi_t \mid s_t)
    \pi_{\theta}(\theta_t \mid s_t,\phi_t)
    \pi_{v}(v_t \mid s_t,\phi_t,\theta_t) \nonumber \\
    &\quad \times
    \pi_{\mathrm{rec}}(r_t \mid s_t,\phi_t,\theta_t,v_t).
    \label{eq:policy-factorization}
\end{align}
The first three factors define the shot geometry and speed, and the final factor chooses post-shot recovery.
On receiver turns, the same encoder instead produces the masked categorical distribution $\pi_{\mathrm{int}}(I_t \mid s_t,a_t^{\mathrm{in}},\mathcal{M}_t)$ over the candidate interception points, where $a_t^{\mathrm{in}}$ is the incoming shot and $\mathcal{M}_t$ is its feasible-interception mask.
The receiver and hitter branches are selected on consecutive role-specific simulator steps rather than sampled as one joint action.
This is illustrated in Figure~\ref{fig:overview}.
This factorization matches the sport's decision structure and allows individual components to be visualized under fixed contact states.
It also permits later action factors to use validity constraints that depend on earlier choices and the current shot geometry: after selecting a direction, the feasible elevation and speed ranges can be narrowed to values that keep the shuttle physically plausible and tactically meaningful from that contact state.
This mirrors the sport's shot-selection process, in which a player first chooses the intended direction and trajectory family, then refines the shot's height, angle, and pace before recovering for the next exchange.
The discrete factorization is chosen deliberately over continuous control in the analyzed experiments.
Continuous launch and recovery parameters would make the action space more expressive, but the binned representation improves training stability, keeps invalid-action handling tractable, and makes the learned strategy inspectable as a distribution over named shot and recovery choices.
The resulting tactical richness should therefore be read as variation within this interpretable discretized control interface, not as evidence that the same policy would be unchanged under a continuous-action formulation.
\rev{The experiments do not include a direct factorized-versus-monolithic policy ablation, so we make no claim that factorization universally improves competitive performance.
}

\subsection{PPO Self-Play}
Training uses PPO with parallelized environments.
The current policy plays against opponents sampled from a checkpoint pool.
To reduce forgetting while keeping pressure from recent policies, opponent sampling is recency-biased.
A small probability of heuristic opponents is retained to anchor basic play and avoid collapse.
The main training schedule first runs to $3.0$M self-play timesteps with pure recency-biased sampling from the continuation checkpoint pool.
Pure-recency evaluation tends to plateau near this point, consistent with over-emphasis on the newest opponents and possible overfitting to a narrow local style.
The $3.0$M--$6.0$M continuation therefore uses a broader linear-recency scenario: most opponent samples are drawn from historical anchors with linear recency weighting, while a smaller fraction is reserved for recent continuation checkpoints, the newest continuation checkpoint, and heuristic opponents.
\rev{The evaluation section compares these regimes across five independent seeds; the detailed matchup matrix retains one lineage to show matchup-specific structure.}
The configuration used for the run in this paper is summarized in Table~\ref{tab:runconfig}.

\begin{table*}[t]
\centering
\small
\begin{tabular}{p{0.22\textwidth}p{0.72\textwidth}}
\toprule
Component & Setting \\
\midrule
Simulator & Singles court; 2D players; 3D shuttle flight with drag-square dynamics. \\
Movement & Speed $5.0$ m/s; acceleration $8.0$ m/s$^2$; racket length $1.6$ m; reaction time $0.15$ s. \\
Action space & Receiver: up to 20 masked interception candidates. Hitter: $11$ azimuth $\times$ $8$ elevation $\times$ $5$ speed; $5 \times 5$ recovery grid. \\
Training & PPO self-play; $0$--$3.0$M pure-recency stage followed by $3.0$M--$6.0$M linear-recency continuation; $8$ parallel environments. \\
Opponent pool & Stage 1: $6$ checkpoint opponents with pure recency-biased sampling and $0.05$ heuristic opponent probability. Stage 2: variety pool with $0.70$ historical anchors sampled by linear recency, $0.15$ recent continuation checkpoints, $0.05$ newest continuation checkpoint, and $0.10$ heuristic opponents. \\
Reward & Sparse rally outcome; dense tactical shaping disabled. \\
Recovery update & Factor-specific PPO update; recovery advantage augmented by Counterfactual Recovery Advantage (CRA) with coefficient $0.05$; $24$ alternative recovery samples; one opponent-response sample. \\
Evaluation & Frozen checkpoint evaluation; fixed-pool / round-robin win rates; $200$ rallies per cell. \\
\bottomrule
\end{tabular}
\caption{Key self-play configuration used for the models analyzed in this paper. Results are reported for checkpoints up to $6.0$M self-play timesteps.}
\label{tab:runconfig}
\end{table*}

\paragraph{Recovery-head credit assignment.}
The shot factors use the standard PPO/GAE advantage, while the recovery factor uses a recovery-specific PPO advantage augmented by Counterfactual Recovery Advantage (CRA).
CRA compares the selected recovery with alternative recovery cells under the same shot and opponent-response context, giving the recovery head a targeted credit signal rather than assigning it only the scalar rally advantage.
Appendix~A gives the full CRA formulation and ablation.

\section{Evaluation and Results}

Self-play policies cannot be fully evaluated by a single scalar metric.
A policy can improve against the current pool while becoming worse against an older style, and a local metric such as average shot speed can saturate before tactical behavior stops changing.
Accordingly, frozen checkpoints are evaluated using both competitive and controlled probes.
The evaluation centers on two main questions: whether broader opponent sampling changes apparent self-play dynamics, and whether recovery is learned as a tactical action rather than a cosmetic post-shot choice.
A final sanity check compares aggregate rally, recovery, and landing statistics with human singles rallies from ShuttleSet22 \cite{wang2023shuttleset22}.

\subsection{Competitive Evaluation}
Frozen checkpoints are evaluated in round-robin matches against a fixed pool of opponents.
The primary visualization is a win-rate matrix in which rows are evaluated checkpoints and columns are fixed-pool opponents.
Each cell reports the row policy's rally win rate over $200$ rallies.
For a single 200-rally cell, binomial uncertainty is still visible: near a 50\% win rate, a normal 95\% interval is roughly $\pm 7$ percentage points.

\begin{figure}[!t]
    \centering
    \includegraphics[width=\columnwidth]{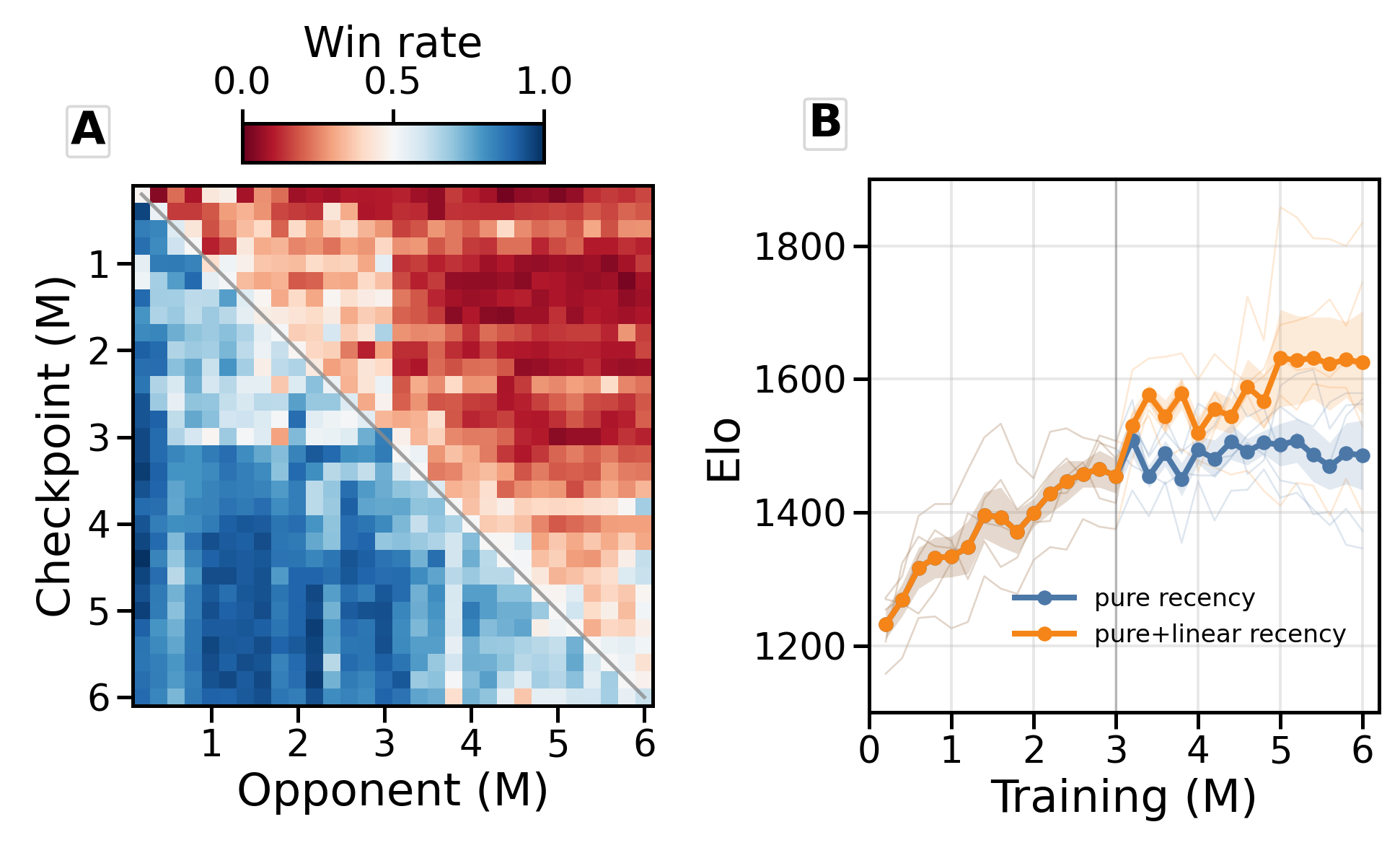}
    \caption{\rev{Competitive evaluation across frozen checkpoints. (A) Illustrative single-lineage win-rate matrix where rows are evaluated checkpoints and columns are 30 fixed-pool opponents sampled every 0.2M timesteps from 0.2M to 6.0M; each cell uses 200 rallies. Off-diagonal patches show style-dependent matchups and possible non-transitivity. (B) Five-seed cross-fit Bradley--Terry/Elo comparison of pure-recency and pure+linear-recency training. Faint lines show individual seeds, thick lines show the mean, shaded bands show standard error, and the vertical line marks the 3.0M-step branch. The shared prefix and post-branch policies are connected through common fixed-pool opponents.}}
    \label{fig:competitive}
\end{figure}

\rev{Figure~\ref{fig:competitive}A shows an illustrative single-lineage round-robin win-rate matrix.}
Later checkpoints usually dominate earlier checkpoints, but the matrix contains off-diagonal structure rather than a perfectly monotonic ordering.
The matrix is therefore more informative than a single average score.
\rev{This is expected in non-transitive self-play populations \cite{lanctot2017unified}: different checkpoints may represent different tactical styles, and a policy can be strong against recent opponents while remaining vulnerable to an older style.}

To quantify the ranking of different checkpoints, the evaluation also fits a fixed-pool Bradley--Terry/Elo-style rating curve.
For two policies $i$ and $j$ with scalar ratings $\rho_i$ and $\rho_j$, the Bradley--Terry win probability is
\begin{equation}
    P(i \succ j) = \frac{\exp(\rho_i)}{\exp(\rho_i)+\exp(\rho_j)}.
\end{equation}
The resulting ratings are used only as a retrospective summary of the win-rate matrix, not as a training reward.
On the conventional Elo scale, 50- and 100-point rating gaps correspond to expected win probabilities of about 57\% and 64\%, respectively, for the higher-rated policy, but this is an effect-size interpretation rather than a confidence interval.
The Elo curves are therefore interpreted as pooled summaries over many such cells rather than as precise claims about small adjacent-checkpoint differences; the analysis emphasizes large and persistent separations across the opponent pool.
\rev{Figure~\ref{fig:competitive}B provides the regime-level comparison across five independent seeds rather than relying on the illustrative lineage in panel A. The cross-fit Elo analysis combines within-branch pairwise records with shared fixed-pool bridge records, placing both continuations on a connected scale through common opponents. Pure-recency evaluation improves early but saturates around $3.0$M steps, whereas the broader pure+linear-recency continuation continues to improve before saturating later, around $5$--$6$M steps. Individual seeds remain variable, but the aggregate separation supports the conclusion that broader opponent sampling delays apparent self-play saturation in this setup.}
The boundary near $3.0$M in the win-rate matrix is consistent with this training-regime shift: after the switch, later policies---now trained against a more varied opponent distribution rather than only the newest local opponent style---obtain high win rates against many earlier pure-recency checkpoints.

\subsection{Controlled Contact Probes}
Badminton shot value is highly contextual: the same trajectory can be attacking or vulnerable depending on contact state, opponent position and velocity, and the hitter's recovery.
The controlled-contact probe therefore fixes the local tactical state and samples shot trajectories from checkpoints across training.
The panels below are representative scenarios.
Appendix~F aggregates these probes across fixed tactical states.

\begin{figure*}[htbp]
    \centering
    \includegraphics[width=\textwidth]{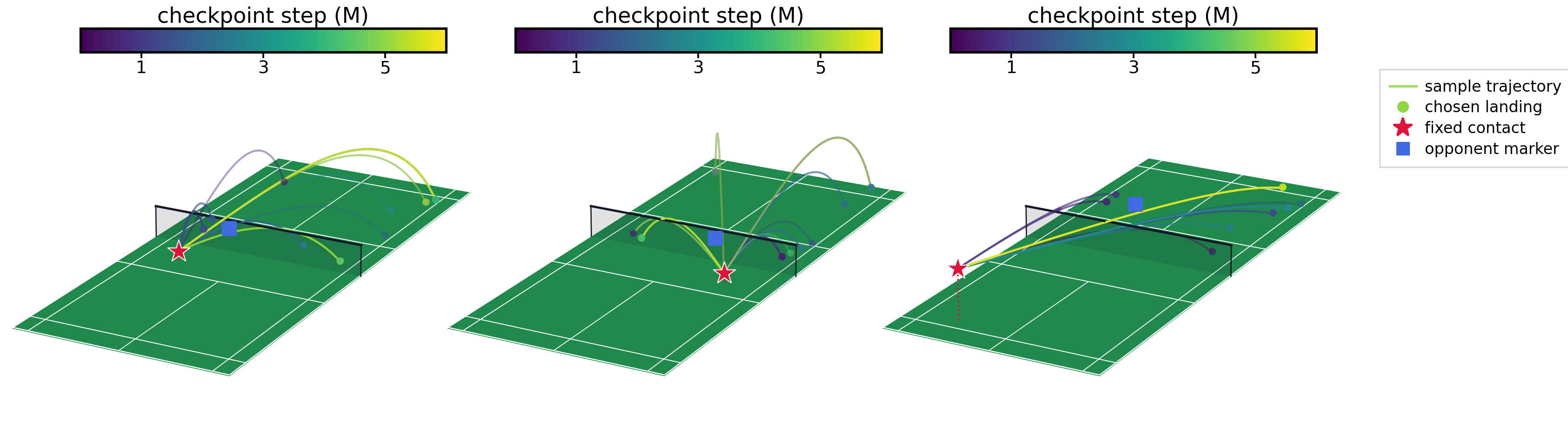}
    \caption{Controlled-contact trajectory samples across checkpoints. Each panel fixes a contact situation and overlays sampled 3D shuttle trajectories, with color indicating checkpoint step. The red star marks the fixed contact state, green markers show sampled landing choices, and the blue square marks the opponent position. Holding the contact state fixed makes changes in shot geometry and landing distribution visible across training.}
    \label{fig:controlled-contact}
\end{figure*}

Figure~\ref{fig:controlled-contact} overlays sampled trajectories for low frontcourt contacts on both sides and a high left-backcourt contact against a midcourt opponent.
Because the inputs are fixed, the changing trajectories reflect changes in the conditional policy rather than changing state visitation.
The shifts are scenario-specific, not merely global changes in average speed or depth.

\begin{figure*}[htbp]
    \centering
    \includegraphics[width=\textwidth]{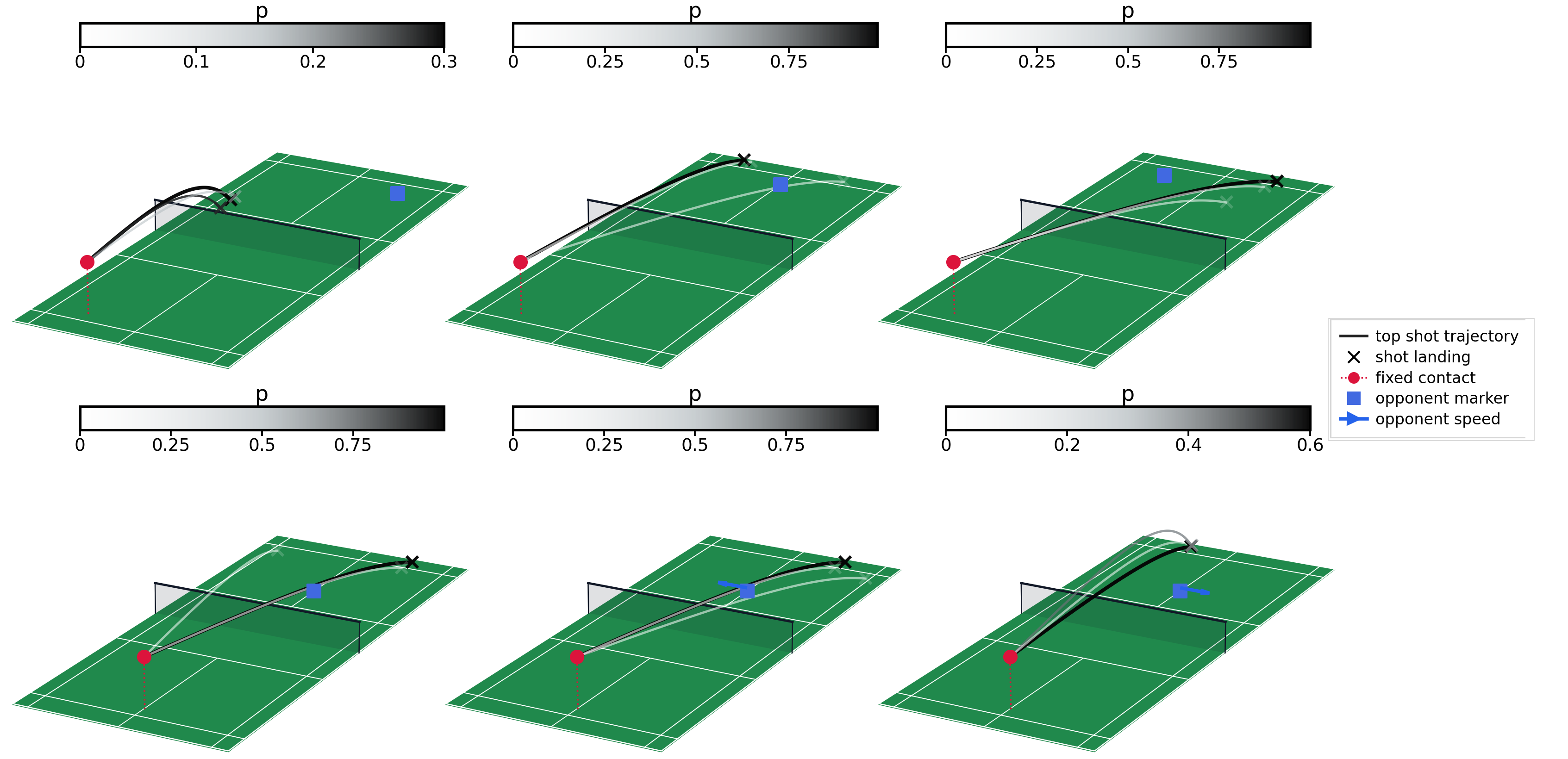}
    \caption{Top-three shot trajectories for two fixed backcourt tactical states. The top row fixes a high left-backcourt contact while the opponent is positioned in the right backcourt; the bottom row fixes a high central-backcourt contact while the opponent starts in midcourt and moves laterally at $5\,\mathrm{m/s}$, indicated by the blue arrow. Each panel shows one high-probability shot mode; grayscale indicates action probability, the red marker shows the fixed contact location, the blue square shows the opponent position, and the black cross marks the projected landing point. The high-probability modes differ in landing depth, cross-court direction, speed, and trajectory geometry, showing how the policy conditions shot selection on opponent location and velocity.}
    \label{fig:top-shots-backcourt}
\end{figure*}

Figure~\ref{fig:top-shots-backcourt} shows the top-three shot modes for two fixed backcourt states.
When the opponent is displaced or moving away from the target side, fast deep trajectories become high-probability options; when the opponent is already waiting in the backcourt, softer and shorter alternatives receive more probability.
The high-probability modes differ in direction, depth, speed, and trajectory geometry, indicating that the policy conditions on opponent state rather than repeating a universal shot preference.

\begin{figure*}[h]
    \centering
    \includegraphics[width=\textwidth]{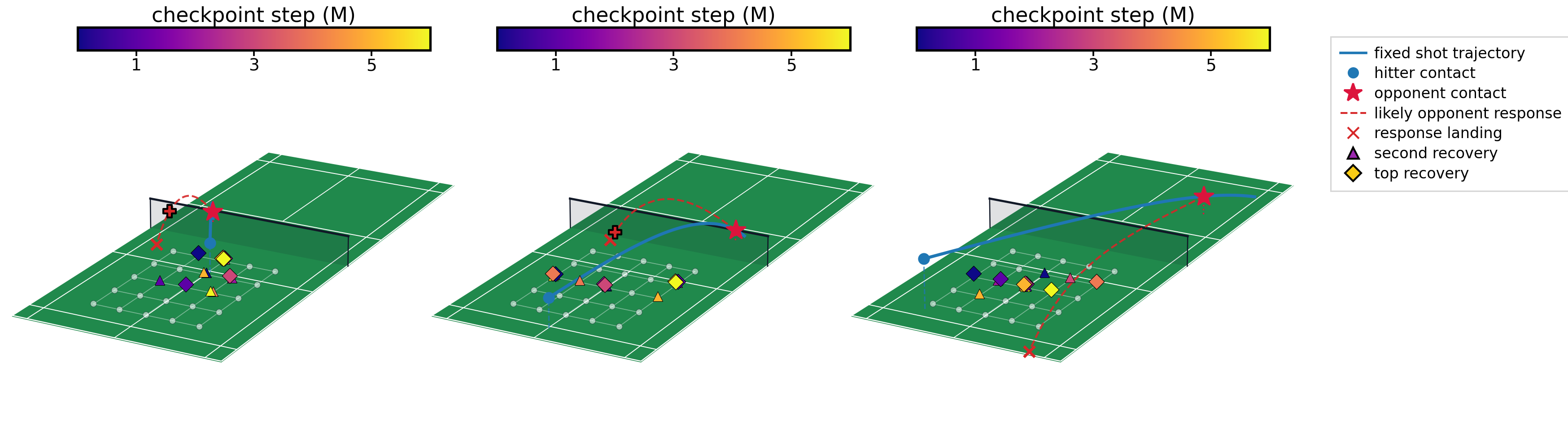}
    \caption{Evolution of top recovery choices for fixed shot probes. The blue curve shows the fixed outgoing shot trajectory, the blue dot marks hitter contact, the red star marks expected opponent contact, and the dashed red curve shows a likely opponent response. Colored diamonds and triangles show the top and second recovery choices across checkpoints, revealing how recovery targets adapt to the same controlled tactical situation during training.}
    \label{fig:recovery-evolution}
\end{figure*}

\subsection{Recovery Evolution Probes}

Recovery is evaluated with the same fixed-context logic.
The probe fixes a shot and a sampled opponent response, then visualizes the top recovery choices at different checkpoints under that same shot-response context.
This separates changes in the recovery policy from changes in the distribution of shots and rallies encountered during self-play.

Figure~\ref{fig:recovery-evolution} shows that recovery choices change under fixed shot-response contexts.
The learned policy does not simply return to court center; it discovers context-dependent targets, such as following a short shot toward the net to cover the immediate net reply.
Thus the probe answers the second question: recovery is an actively learned tactical decision, not an afterthought attached to shot selection.
This is the role of CRA in training: it provides a targeted credit signal that helps separate the recovery choice from the value of the shot itself.

In addition to the quantitative and controlled-probe evaluations, full rally rollouts provide qualitative inspection of the learned behavior.
Appendix~B provides representative rally rollouts.

\subsection{Human-Data Sanity Check}

\begin{figure*}[htbp]
    \centering
    \includegraphics[width=\textwidth]{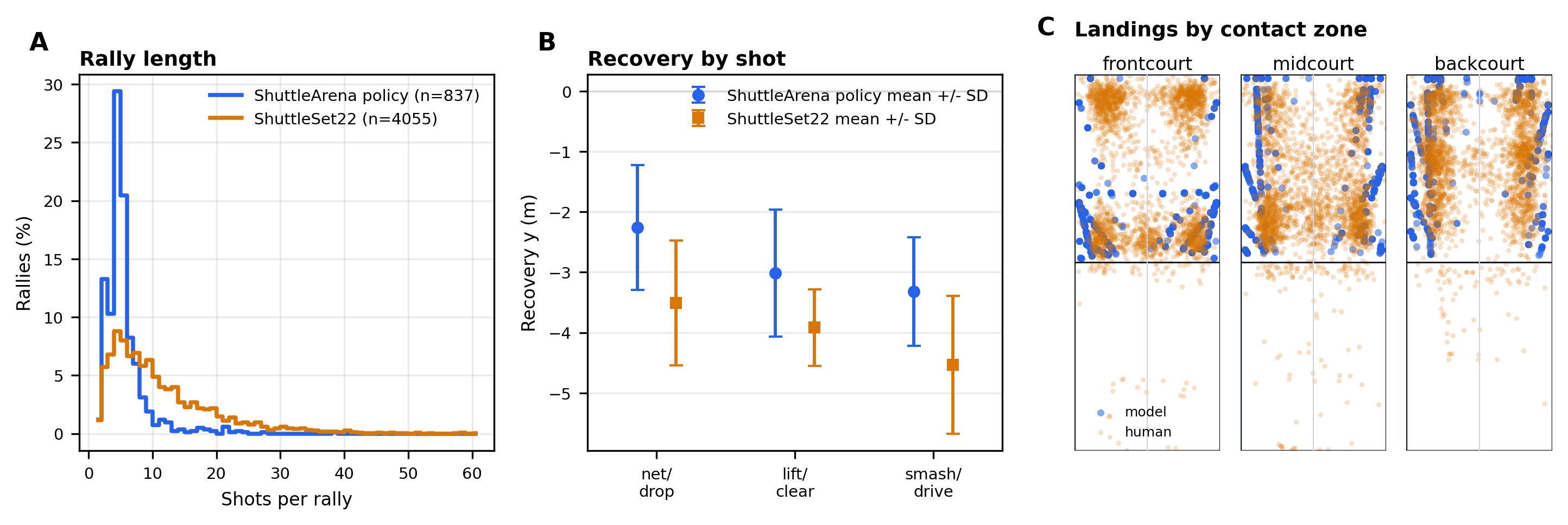}
    \caption{Sanity comparison between ShuttleArena policy rollouts and human singles rallies from ShuttleSet22. (A) Rally-length distributions across different seeds show that the learned policy produces shorter rallies than the human dataset. (B) Recovery targets are shallower, especially after net/drop, lift/clear, and smash/drive groups. (C) Landing distributions by contact zone are broadly similar in court regions, but the learned policy places more shots near sidelines and uses fewer net-contact lifts.}
    \label{fig:human-sanity}
\end{figure*}

Figure~\ref{fig:human-sanity} compares aggregate behavior from learned ShuttleArena rollouts with human rallies from ShuttleSet22.
\rev{The ShuttleSet22 subset represents professional international singles play from major BWF tournaments; after filtering to non-flawed singles rallies, our subset contains 43{,}994 stroke events from 4{,}055 rallies across 140 set files, drawn from source metadata covering 58 matches and 35 players. This comparison is not an imitation target, an external-policy comparison, or a calibrated realism score. It is a human-data sanity check on spatial and rally-level behavior, and it does not establish that ShuttleArena reproduces the full distribution of professional match play.}

Panel A shows that ShuttleArena rallies are shorter than the ShuttleSet22 rallies.
\rev{
Plausible contributors include simplified player motion and interception, exact execution without human motor or perceptual noise and fatigue, simulator-specific tactical exploitation, randomized rally initialization rather than full match context, and optimization for rally reward rather than human imitation.
}
Panel B shows that the learned recovery targets are systematically closer to the front court.
Human players may also intend to recover forward after short or attacking shots, but their actual recovery positions are shaped by continuous transition into the next running action and the opponent's reply rather than by a single discrete target; in the simulator, the selected recovery target is represented more directly.
Panel C shows broadly similar landing structure across frontcourt, midcourt, and backcourt contacts, while also revealing sharper model concentration near the sidelines.
This difference is consistent with the machine-controlled policy being able to aim more precisely at narrow court regions.
The frontcourt panel also shows fewer learned lifts from net contacts, which is tactically plausible because high lifts from the net can give the opponent an attacking opportunity.
Overall, the comparison supports the qualitative interpretation that the learned policy discovers recognizable badminton-like placement and recovery patterns, while also reflecting the precision and abstraction of the simulated environment.

\section{Ablations}

\begin{figure}[htbp]
    \centering
    \includegraphics[width=\columnwidth]{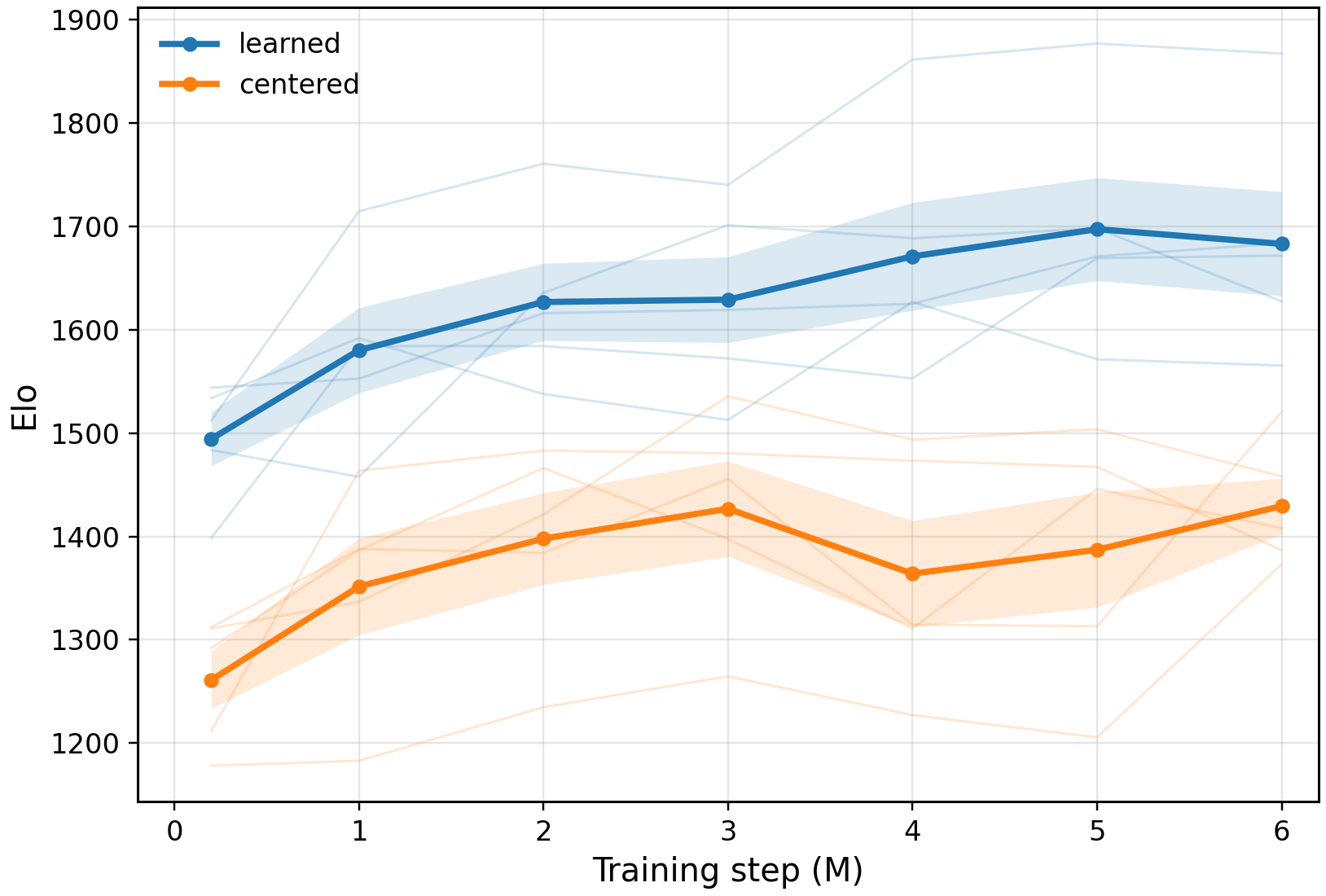}
    \caption{\rev{Recovery-choice intervention across five independent seeds. The shot policy is held fixed for each checkpoint, while the recovery factor is either the learned policy or a centered override. Lines show mean Elo-style ratings and shaded bands show standard error across seeds. The separation measures the total competitive consequence of replacing learned recovery, including the downstream state-distribution shift.}}
    \label{fig:ablations}
\end{figure}

The main ablation is an evaluation-time intervention on recovery execution.
For each checkpoint in each of five independent seeds, the learned shot policy is held fixed while the recovery factor is evaluated either as the learned recovery policy or as a centered recovery override selected from the same feasible recovery grid.
The learned-recovery and centered-recovery variants are evaluated against shared fixed opponent pools, and the resulting pairwise outcomes are summarized by the same Bradley--Terry/Elo procedure used above.

\rev{Figure~\ref{fig:ablations} shows that learned recovery behavior is competitively important. At 6.0M steps, the learned-recovery variant reaches approximately 1683 Elo on average, while the centered-recovery variant reaches approximately 1429 Elo. Although the shot policy is held fixed at the intervention point, the centered target changes both immediate court coverage and the downstream state distribution. The roughly 250-Elo gap therefore measures the total competitive consequence of the intervention; it is not a local causal decomposition in which future states are otherwise identical.}
Appendix~E tests the learned-policy conclusions under evaluation-time environment perturbations.

\section{Discussion}

The experiments highlight why self-play agents in racket sports need both competitive and behavioral evaluation.
The win-rate matrix exposes non-monotonic matchup structure that a single scalar score can hide, while the retrospective rating curve provides a pooled effect-size summary.

The controlled probes complement this view by fixing tactical inputs and making policy changes inspectable.
Together with the centered-recovery ablation and CRA branch comparison (Appendix~A), they show that recovery is coupled to shot choice and suggest that factor-specific credit assignment can help this part of learning in the tested setup.

\section{Limitations and Future Work}

ShuttleArena intentionally abstracts badminton: players move in two dimensions, shuttle flight is three-dimensional but drag-based, shot and recovery choices are discretized, interception dynamics are lightweight, and CRA uses a single sampled opponent-response context.
The model does not yet include full biomechanics, deceptive preparation, fatigue, perception uncertainty, spin-dependent aerodynamics, or detailed racket contact mechanics.
These choices keep self-play fast while making the shot--recovery coupling inspectable, but they also mean that some tactical structure may reflect the chosen action parameterization and simplified opponent-response dynamics.
The present results establish that this rally-level abstraction supports competitive self-play, interpretable tactical change, and useful recovery credit assignment; testing which patterns survive continuous control, richer player kinematics, multi-sample opponent responses, and calibration from human match trajectories is the next fidelity step.
\rev{The human comparison is also limited in scope: it covers a filtered professional-singles population and exposes a clear rally-length gap, but it does not validate human-equivalent dynamics or generalization to recreational players, doubles, full scored matches, or populations outside those tournaments.}




Several extensions follow naturally: style-controllable agents, human-versus-computer play, doubles with partner coverage and rotation, and calibration from human match trajectories, stroke annotations, and recovery patterns.

\section{Reproducibility Notes}

The analyzed run uses the self-play parameters in Table~\ref{tab:runconfig}.
Appendix~C lists the detailed network and PPO hyperparameters.
Key reproducibility details include the checkpoint-pool sampling rule, sparse reward setting, disabled curriculum, physics and movement constants, action discretization, and factor-specific recovery PPO update with CRA coefficient $0.05$, $24$ alternative recovery samples, one opponent-response sample, and no distribution auxiliary.
The ShuttleArena code and reproducibility materials are available at \projecturl.

\section{Conclusion}

ShuttleArena is a physics-based badminton self-play environment in which shot selection and post-shot recovery are learned as coupled tactical decisions.
Across frozen-checkpoint evaluations, later policies improve substantially while still showing non-monotonic, matchup-dependent dynamics.
Controlled probes show scenario-specific changes in shot geometry and recovery choice, and ablations show that recovery is a meaningful tactical action whose training can benefit from targeted credit assignment.
The results suggest that physics-based racket sports are a compact and interpretable testbed for game AI, where self-play improvement can be studied through both win rates and visible tactical change.

\ifwithappendix
\section{Appendix A: Counterfactual Recovery Advantage}

\subsection{CRA Algorithm}
A standard PPO update assigns the same scalar advantage to all components of an action.
For badminton recovery, this is noisy: if a rally is won, the recovery target may receive positive credit even when the win was caused by the shot itself; if a rally is lost, the recovery target may be blamed even when the selected shot was poor or the opponent response was unusually strong.
The structured policy is therefore trained with a recovery-specific advantage for the recovery head.
Counterfactual Recovery Advantage (CRA) starts from the sampled transition that actually occurred in the rollout and adds a comparison against alternative recoveries under the same shot and opponent-response context.

Let $a^{\mathrm{shot}}_t=(\phi_t,\theta_t,v_t)$ be the chosen shot, $r_t$ the selected recovery target, and $s_{t+1}$ the next observation after the simulator advances the rally.
The critic estimates the value before the contact:
\begin{equation}
    v_t
    =
    V_{\psi}(s_t).
\end{equation}
For nonterminal recovery transitions, the learning target is the critic value of the next observation.
For terminal transitions with an explicit rally outcome, the learning target is the terminal reward:
\begin{equation}
    y^{\mathrm{rec}}_t
    =
    \begin{cases}
    R_t, & \text{Rally terminates},\\
    V_{\psi}(s_{t+1}), & \text{otherwise}.
    \end{cases}
\end{equation}
The recovery-specific advantage used for the recovery head is
\begin{equation}
    A^{\mathrm{trans}}_{\mathrm{rec},t} = y^{\mathrm{rec}}_t - v_t.
    \label{eq:sampled-recovery-advantage}
\end{equation}

For the counterfactual term, the simulator also evaluates the selected recovery and $24$ alternative recovery cells from the $5\times5$ recovery grid.
Let $\hat{o}_t$ denote the sampled or expected opponent response conditioned on the selected shot.
In the analyzed run, this response context is generated from one simulated opponent-response sample conditioned on the selected shot, and the same response context is reused when comparing recovery targets.
This single-response implementation keeps the counterfactual update inexpensive; using multiple opponent-response samples could reduce estimator variance or bias in the recovery baseline and is a natural extension.
For a candidate recovery target $r$, the simulator constructs the post-shot state
$x_t(r; a^{\mathrm{shot}}_t,\hat{o}_t)$, and the critic assigns it a score
\begin{equation}
    q_{\mathrm{rec}}(r)
    =
    V_{\psi}\!\left(x_t(r; a^{\mathrm{shot}}_t,\hat{o}_t)\right).
\end{equation}
The average score of the sampled alternative recovery targets is used as the baseline:
\begin{equation}
    b_{\mathrm{rec}}
    =
    \frac{1}{|\mathcal{R}'|}
    \sum_{r' \in \mathcal{R}'}
    q_{\mathrm{rec}}(r'),
\end{equation}
where $\mathcal{R}'$ contains the sampled alternatives and excludes the selected recovery.
The CRA term is
\begin{equation}
    A^{\mathrm{cf}}_{\mathrm{rec},t}
    =
    q_{\mathrm{rec}}(r_t) - b_{\mathrm{rec}}.
    \label{eq:counterfactual-recovery-advantage}
\end{equation}
The advantage passed to the recovery PPO surrogate is
\begin{equation}
    \tilde{A}^{\mathrm{rec}}_t
    =
    A^{\mathrm{trans}}_{\mathrm{rec},t}
    +
    \lambda_{\mathrm{cf}} A^{\mathrm{cf}}_{\mathrm{rec},t},
    \qquad
    \lambda_{\mathrm{cf}}=0.05.
    \label{eq:combined-recovery-advantage}
\end{equation}
This combined advantage is only applied to transitions in which the policy selected a recovery factor.
It is normalized over active recovery transitions in the PPO minibatch.

The shot factors still use the standard PPO/GAE advantage $A_t$.
The recovery factor instead uses $\tilde{A}^{\mathrm{rec}}_t$ inside a clipped PPO surrogate:
\begin{align}
    \rho^{\mathrm{rec}}_t
    &=
    \frac{
    \pi_{\mathrm{rec}}(r_t \mid s_t,a^{\mathrm{shot}}_t)
    }{
    \pi^{\mathrm{old}}_{\mathrm{rec}}(r_t \mid s_t,a^{\mathrm{shot}}_t)
    },
    \\
    \mathcal{L}_{\mathrm{rec}}
    &=
    -\mathbb{E}_t\!\left[
    \min\!\left(
    \rho^{\mathrm{rec}}_t \tilde{A}^{\mathrm{rec}}_t,
    \right.\right. \nonumber \\
    &\qquad\left.\left.
    \mathrm{clip}(\rho^{\mathrm{rec}}_t,1-\epsilon,1+\epsilon)
    \tilde{A}^{\mathrm{rec}}_t
    \right)
    \right].
    \label{eq:recovery-ppo-loss}
\end{align}
The policy loss is the sum of the shot-factor PPO loss and this recovery-factor PPO loss, with the usual value and entropy terms.
Thus the CRA term is not a separate supervised loss; it changes the recovery advantage used inside the PPO recovery surrogate.
Operationally, each PPO iteration samples opponents from the configured checkpoint pool, rolls out parallel rallies, computes standard PPO/GAE advantages for the shot factors, computes $\tilde{A}^{\mathrm{rec}}_t$ for hitter contacts with selected recovery targets, updates the shot and recovery factors with their corresponding PPO surrogates, and periodically inserts the current policy into the opponent pool.

\subsection{CRA Ablation}
The CRA ablation compares two independently trained lineages: the default run uses CRA with 24 alternative recovery samples and coefficient $0.05$, while the ablated run disables CRA by using zero alternative recovery samples and coefficient $0$.
A cross-play experiment freezes checkpoints from both lineages, evaluates them against a common mixed opponent pool drawn from both lineages, and fits one Elo scale to all pairwise matches.

\begin{figure}[h]
    \centering
    \includegraphics[width=\columnwidth]{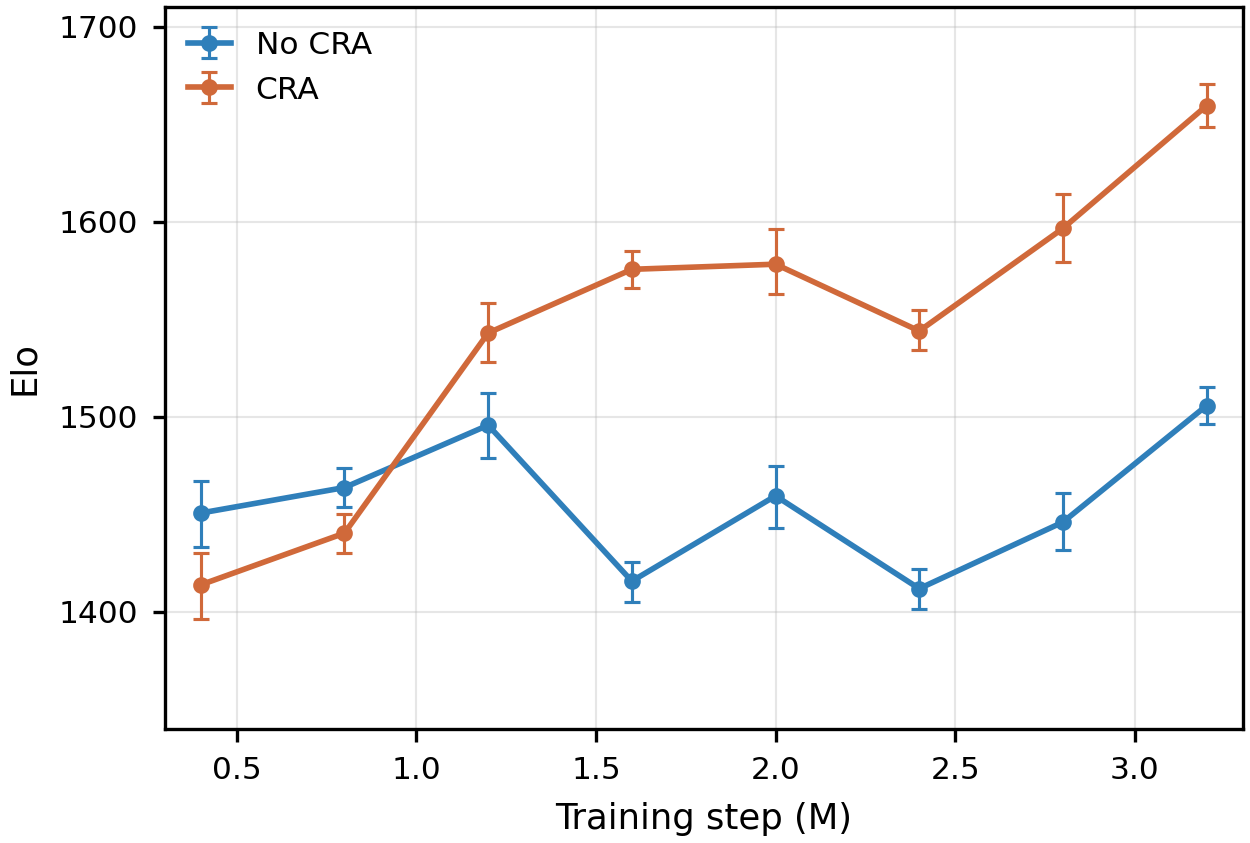}
    \caption{CRA/no-CRA training ablation for the main lineages. Checkpoints from the CRA and no-CRA lineages are evaluated against a common mixed fixed pool with 200 rallies per pair and summarized with one Bradley--Terry/Elo fit. Error bars show 95\% bootstrap intervals obtained by resampling pair outcomes and refitting Elo. The CRA run is rated higher after early training in this evaluation.}
    \label{fig:appendix-cra-ablation}
\end{figure}

Figure~\ref{fig:appendix-cra-ablation} shows that the CRA run is rated higher than the matched no-CRA run after early training; at 3.2M steps, the default run is about 1660 Elo versus about 1506 Elo without CRA.
This 154-Elo separation is based on all 200-rally cells against the shared pool, not on a single pairwise matchup.
As a simple uncertainty check, the six common-opponent evaluation cells at 3.2M are bootstrapped; the CRA run remains above the ablated run in 100\% of 1{,}000 paired bootstrap resamples, with a 95\% bootstrap interval of roughly 9--25 percentage points for the mean win-rate advantage.
Overall, this earlier lineage-level comparison suggests that the CRA recovery-head update can help in this setup.

\section{Appendix B: Qualitative Rally Rollouts}

Qualitative rally visualizations complement the scalar competitive summaries and controlled tactical probes.
They are not used as additional scores; instead, they provide a direct check that the learned policies produce coherent multi-shot exchanges.
Matched rollouts are rendered from frozen checkpoints against the same fixed opponent and initial seeds, allowing visual comparison of shot execution, recovery movement, interception timing, and failure modes.

Figure~\ref{fig:rally-exhibition} shows example rallies rendered as stage-by-stage court snapshots.
The examples illustrate the core ShuttleArena loop: hitter contact, outgoing shuttle flight, receiver interception or miss, hitter recovery, and continuation or termination.
The figure supports qualitative inspection rather than statistical ranking.
It shows that ShuttleArena produces readable rally sequences, illustrates behavioral differences between early and late policies, and makes learned decisions visible at the level of individual exchanges.
For a direct comparison, the 0.2M-step and 6.0M-step checkpoints are evaluated against the same 3.0M-step opponent.
The first row shows the 0.2M-step policy losing a rally after a loose drop shot.
The early policy produces a readable exchange, but the drop leaves the shuttle attackable and exposes a poor tactical choice rather than a controlled construction of the point.
The remaining two rows show the 6.0M-step policy in both attacking and defensive situations.
In the first 6.0M example, the agent uses a direct attacking pattern: it creates an opening and finishes with a cross-court smash, a shot choice that is qualitatively consistent with badminton tactics because it changes both depth and lateral direction at high speed.
In the second 6.0M example, the agent uses a more extended sequence around the net.
It plays a cross-court net drop, responds to the opponent by lifting high, absorbs the opponent's smash, and then counterattacks by driving the shuttle quickly into the opposite corner.
This final transition is informative qualitatively: instead of treating defense as only passive survival, the learned policy converts an opponent attack into a fast counterattack, a pattern consistent with common badminton tactics in which a player withstands pressure and then redirects the rally into the open court.

\begin{figure*}[htbp]
    \centering
    \includegraphics[width=\textwidth]{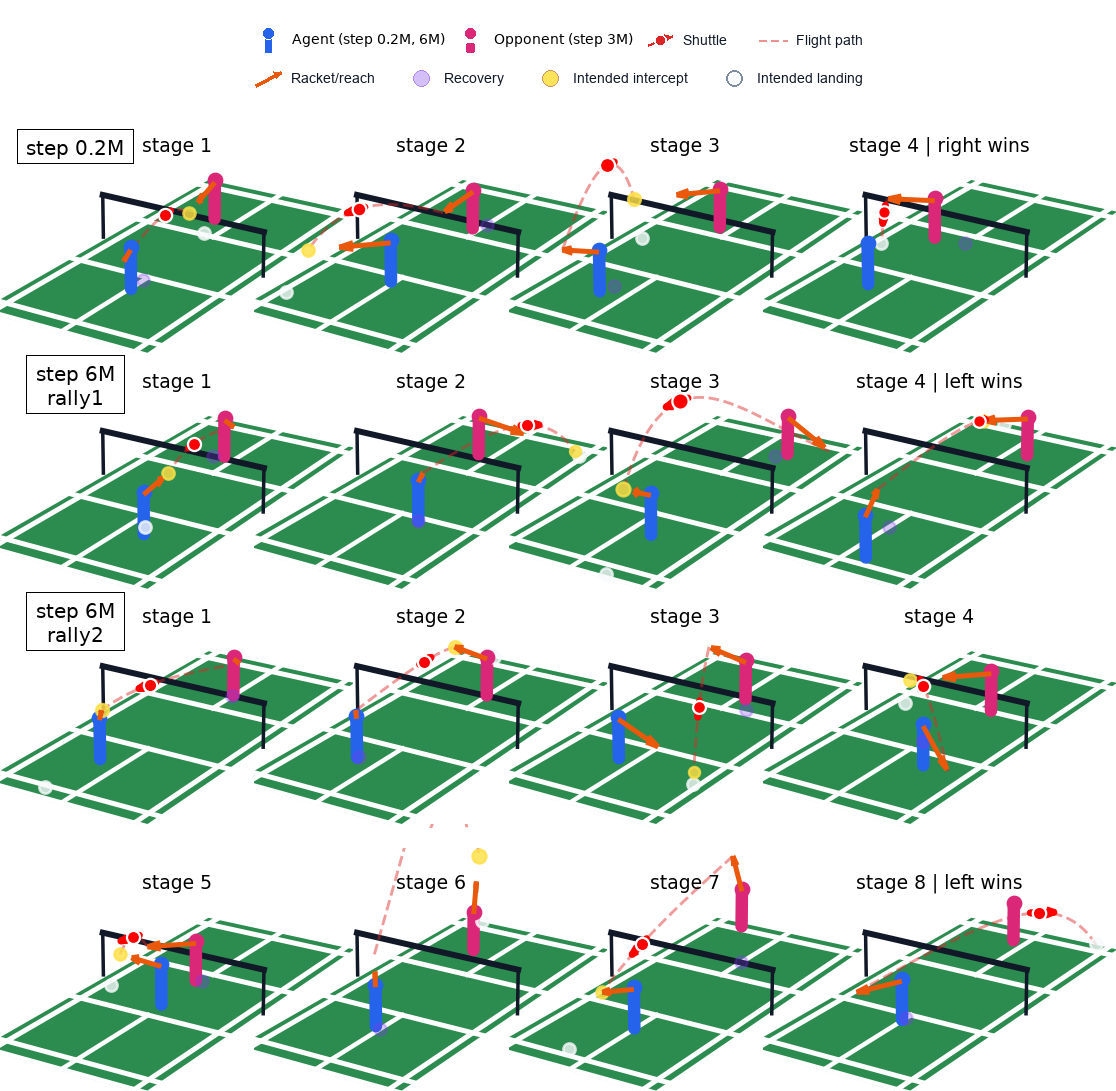}
    \caption{Example ShuttleArena rallies rendered as stage-by-stage court snapshots. Each panel shows the active players, shuttle, outgoing flight path, intended intercept or landing point, and selected recovery target. The examples illustrate how ShuttleArena couples shot execution, receiver interception, and post-shot recovery across multi-stage exchanges with different terminal outcomes.}
    \label{fig:rally-exhibition}
\end{figure*}

An illustrative video is available in the project GitHub repository at \projecturl.

\section{Appendix C: Detailed Training Parameters}

Table~\ref{tab:ppohparams} reports the network and PPO settings used for the analyzed ShuttleArena self-play run.
The policy and value networks use a small two-layer MLP with a factorized hitter branch and a separate masked receiver-interception head, while the optimization settings follow a standard PPO configuration with clipped policy updates, generalized advantage estimation, entropy regularization, and minibatch training over parallel rollouts.
These values are included to make the reported training runs easier to reproduce and to distinguish algorithmic choices such as CRA and opponent-pool sampling from ordinary PPO hyperparameters.

\begin{table}[h]
\centering
\small
\begin{tabular}{p{0.44\columnwidth}p{0.46\columnwidth}}
\toprule
Hyperparameter & Value \\
\midrule
Hidden layers & Policy/value MLP $[64,64]$; factorized hitter heads; separate receiver head \\
Activation & Tanh \\
Learning rate & $3\times10^{-4}$ \\
PPO clip $\epsilon$ & $0.2$ \\
Discount $\gamma$ & $0.99$ \\
GAE $\lambda$ & $0.95$ \\
Rollout length & $256$ steps/env; $8$ envs \\
Batch / minibatch & $2048$ rollout transitions; minibatch $256$ \\
PPO epochs & $10$ \\
Entropy coefficient & $0.002$ \\
Value coefficient & $0.5$ \\
Optimizer & Adam \\
Checkpoint cadence & $2000$ timesteps \\
Random seed & $17$ \\
\bottomrule
\end{tabular}
\caption{Network and PPO hyperparameters for the primary analyzed self-play run. Values are from the run configuration or the Stable-Baselines3 defaults used when the training script does not override them.}
\label{tab:ppohparams}
\end{table}

\section{Appendix D: Shuttle Trajectory Validation}

The main environment combines simplified two-dimensional player motion with three-dimensional shuttle flight under drag-square dynamics.
This section evaluates whether the chosen shuttle parameters produce badminton-like trajectories for representative shot families.
The validation is limited to the shuttle-flight layer and is not intended as a full model of biomechanics or racket contact.
The simulated trajectories are compared with external reference quantities for high clears/lifts, drops, smashes, and drives.

Figure~\ref{fig:appendix-shuttle-validation} compares simulated side-view trajectories to shot-specific reference bands.
For the clear/lift, the reference is the BWF shuttle-speed test: a correct-speed shuttle, hit with a full underhand stroke from the back boundary line, should land 0.53--0.99 m short of the opposite back boundary line \cite{bwflaws2026}.
For the smash, the reference is the measured high-speed-camera analysis of \citet{collet2026shuttlecock}, which reports approximately exponential velocity decay, a velocity-halving distance near 3.35 m, and a roughly 0.44 s time-of-flight over 10 m for a 300 km/h smash.
For drops and drives, public quantitative trajectory datasets are less standardized, so the checks use conservative geometry and speed criteria: drops should land close to the net at low terminal speed, and drives should remain low and flat while clearing the net.
The simulated examples satisfy all baseline checks used to generate the figure: the clear/lift lands 0.74 m short of the far back line; the smash reaches 10 m in 0.40 s and has speed ratio $V(3.35\mathrm{m})/V_0=0.51$; the drop lands 1.74 m past the net at 6.4 m/s; and the drive crosses the net at 1.80 m with maximum height 1.84 m.

\begin{figure*}[htbp]
    \centering
    \includegraphics[width=\textwidth]{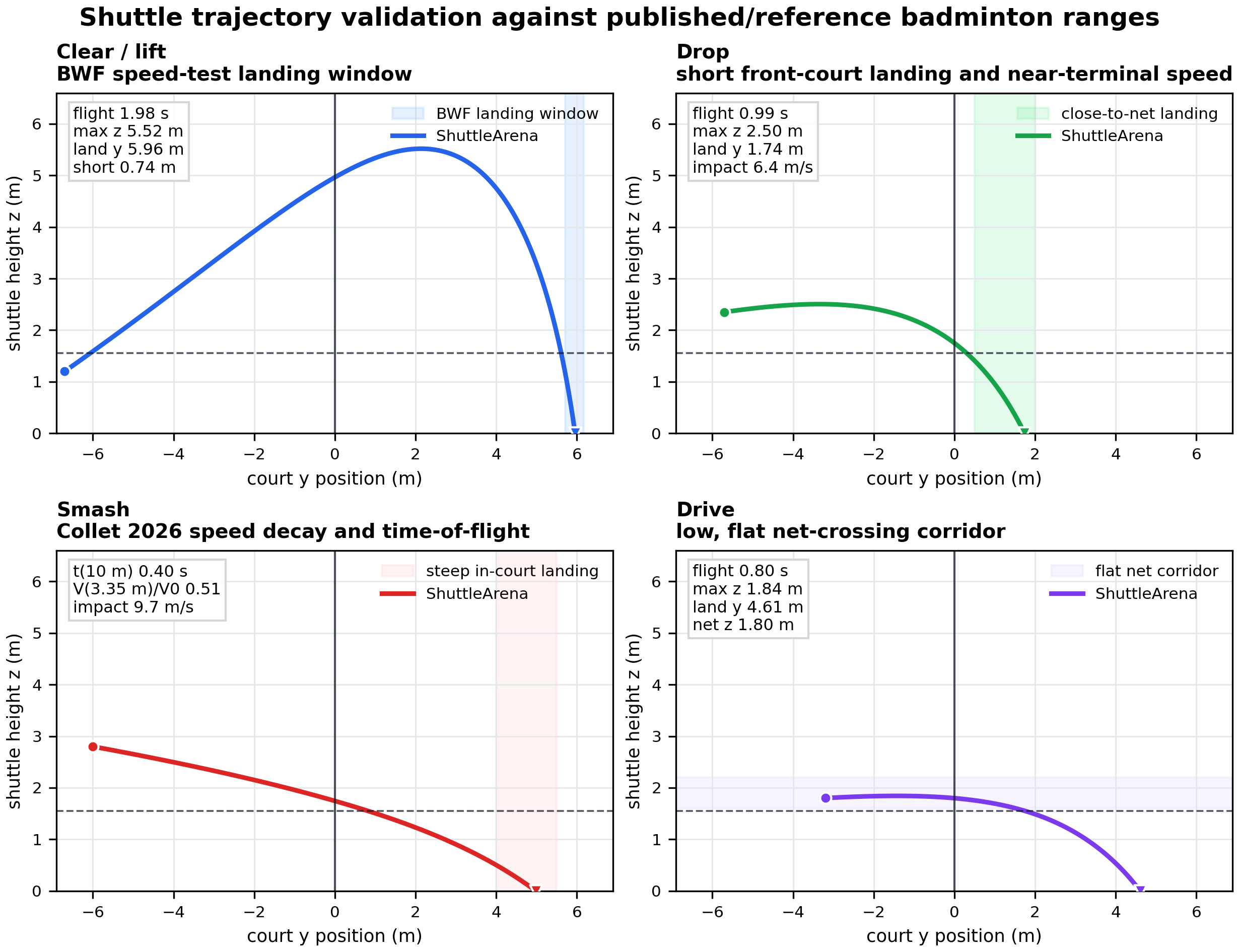}
    \caption{Shuttle trajectory validation for four canonical shot families. Each panel shows the side-view ShuttleArena trajectory under the paper's drag-square parameters ($k_h=0.20$, $k_v=0.16$). Shaded regions indicate external or canonical reference bands: the BWF correct-speed clear/lift landing window, close-to-net drop landing, Collet's smash flight-time and speed-decay range, and a low drive net-crossing corridor. These checks are intended as plausibility validation of the shuttle-flight model, not as detailed calibration of player biomechanics or racket contact.}
    \label{fig:appendix-shuttle-validation}
\end{figure*}

Table~\ref{tab:appendix-shuttle-robustness} reports a simple robustness sweep over the drag coefficients.
Both coefficients are scaled together from $0.8$ to $1.2$ times their nominal values while keeping the same launch examples fixed.
This is a deliberately strict perturbation because the same policy-scale actions are reused without retuning shot launches for each drag setting.
As expected, clear depth is the most sensitive metric, because high clears use most of the court length.
The smash time-to-10 m and drop landing depth change smoothly with drag, and the drive remains low across the sweep.
The nominal setting is also physically interpretable: $k_h=0.20$ gives a horizontal speed-halving distance $\ln 2/k_h=3.47$ m, close to the measured 3.35 m scale reported by \citet{collet2026shuttlecock}, while $k_v=0.16$ gives a vertical terminal speed $\sqrt{g/k_v}=7.83$ m/s, close to published feather-shuttle terminal-speed values.

\begin{table*}[htbp]
\centering
\small
\begin{tabular}{rrrrrrrr}
\toprule
$k$ scale & $k_h$ & $k_v$ & Half-dist. & $V_T$ & Clear short & Drop land & Smash $t_{10}$ \\
\midrule
0.8 & 0.160 & 0.128 & 4.33 & 8.75 & -1.69 & 2.57 & 0.304 \\
0.9 & 0.180 & 0.144 & 3.85 & 8.25 & -0.36 & 2.13 & 0.347 \\
1.0 & 0.200 & 0.160 & 3.47 & 7.83 & 0.74 & 1.74 & 0.398 \\
1.1 & 0.220 & 0.176 & 3.15 & 7.47 & 1.67 & 1.39 & 0.459 \\
1.2 & 0.240 & 0.192 & 2.89 & 7.15 & 2.46 & 1.08 & 0.532 \\
\bottomrule
\end{tabular}
\caption{Drag-robustness sweep for the validation shots. The same four launch examples are reused while scaling both drag coefficients. ``Half-dist.'' is the implied horizontal speed-halving distance $\ln 2/k_h$; $V_T$ is the implied vertical terminal speed $\sqrt{g/k_v}$. Clear shortfall is distance short of the far back line; negative values indicate that the fixed clear launch would travel long under reduced drag.}
\label{tab:appendix-shuttle-robustness}
\end{table*}

\section{Appendix E: Evaluation-Time Environment Robustness}

The main experiments use one nominal environment setting: player speed $5.0\,\mathrm{m/s}$, player acceleration $8.0\,\mathrm{m/s^2}$, racket reach $1.6\,\mathrm{m}$, reaction time $0.15\,\mathrm{s}$, horizontal and vertical drag coefficients $k_h=0.20$ and $k_v=0.16$, and a fast-reaction miss probability of $0.8$ for shuttle flight times below $0.1\,\mathrm{s}$.
The robustness sweep tests whether the main qualitative conclusions persist when these constants are perturbed at evaluation time.
The policies remain frozen; only the simulator constants used during rollouts are changed.
This is a conservative test under model mismatch, not a claim that the same policy would be optimal after retraining in each perturbed environment.

The sweep covers three conclusions from the main text.
First, the late main-run policy at $6.0$M steps is compared with the early $3.0$M policy.
Second, the $6.0$M learned recovery policy is compared with an evaluation-time centered-recovery override that keeps the shot policy fixed but replaces the recovery choice with the nearest feasible centered target.
Third, a CRA-trained checkpoint is compared with a matched no-CRA checkpoint; because the locally available no-CRA lineage has anchors through $3.4$M steps, this comparison uses the $3.4$M checkpoint from each lineage.
Each cell in Table~\ref{tab:appendix-env-robustness} averages 1000 side-balanced rallies, equivalent to five 200-rally evaluation pairs.
The reported yes/no entry indicates whether the first policy in the comparison wins more than half of its rallies; the parenthesized number is the corresponding win rate.

\begin{table*}[htbp]
\centering
\small
\begin{tabular}{llccc}
\toprule
Variant & Change & Late beats early & Learned beats centered & CRA branch higher \\
\midrule
Default & Original settings & Yes (0.865) & Yes (0.794) & Yes (0.721) \\
Lower drag & $k_h,k_v$ $-20\%$ & Yes (0.540) & Yes (0.746) & Yes (0.780) \\
Higher drag & $k_h,k_v$ $+20\%$ & Yes (0.781) & Yes (0.763) & Yes (0.860) \\
Slower player & Speed $-15\%$ & Yes (0.698) & Yes (0.805) & Yes (0.842) \\
Faster player & Speed $+15\%$ & Yes (0.868) & Yes (0.802) & Yes (0.673) \\
Longer reaction & Reaction time $+50\,\mathrm{ms}$ & Yes (0.603) & Yes (0.738) & Yes (0.714) \\
No fast-reaction miss & Fast miss probability $=0$ & Yes (0.963) & Yes (0.754) & Yes (0.786) \\
\bottomrule
\end{tabular}
\caption{Evaluation-time environment robustness for the main learned-policy conclusions. Policies are frozen and only the simulator constants are perturbed during evaluation. A ``yes'' means the first policy in the comparison exceeds a 0.5 rally win rate over 1000 side-balanced rallies, equivalent to five 200-rally evaluation pairs.}
\label{tab:appendix-env-robustness}
\end{table*}

The sweep preserves all three conclusions across the tested perturbations.
The lower-drag late-versus-early comparison is the closest cell, with win rate 0.540, suggesting that this conclusion is directionally robust but that its margin can change under environment mismatch.
The learned-recovery comparison is more stable in this sweep, remaining between 0.738 and 0.805 across variants.
Removing the fast-reaction miss mechanism does not reverse the conclusions: the late policy still strongly beats the early policy, learned recovery remains better than centered recovery, and the CRA-trained checkpoint has the higher rally win rate in the matched no-CRA comparison.
Thus the main qualitative claims do not appear to depend on the exact default drag, movement-speed, reaction-time, or fast-miss constants used by the primary run.

\section{Appendix F: Aggregate Probe Action-Space Coverage}

The representative controlled-contact and recovery panels in the main text show individual tactical situations.
This section aggregates the same probe family across many fixed states to characterize systematic changes in the conditional action distributions.
The analysis is computed from cached frozen-checkpoint probe rollouts only; no additional training is performed.

For the shot probes, each contact state in the controlled-contact grid is fixed, and empirical categorical distributions are estimated over shot type, landing zone, and the joint shot-type--landing-zone choice at each checkpoint.
For the recovery probes, a shot and the sampled opponent response context are fixed, and the empirical distribution over recovery-grid cells is estimated.
For each fixed state or context, the analysis computes the entropy $H$ of the empirical categorical distribution, its effective support $\exp(H)$, and the Jensen--Shannon divergence from the corresponding $0$-step distribution.
Effective support should be read as the approximate number of equally used bins: a value near $1$ is nearly deterministic, a value near $2$ indicates about two comparably used choices, and intermediate values indicate one dominant mode with secondary alternatives.

\begin{figure*}[htbp]
    \centering
    \includegraphics[width=\textwidth]{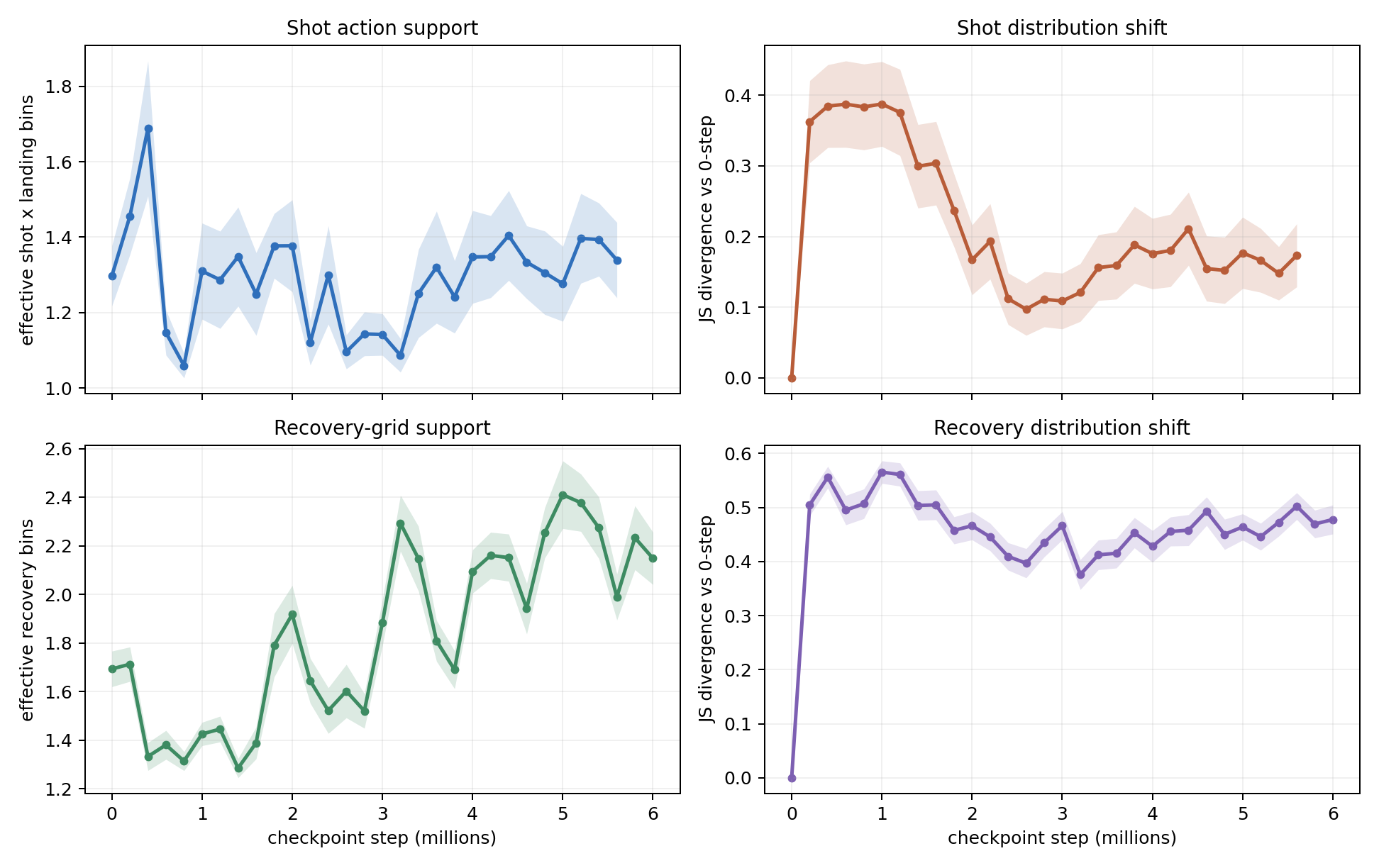}
    \caption{Aggregate controlled-probe action-space coverage and distribution shift. Shot statistics average over $27$ fixed contact states from the controlled-contact grid; recovery statistics average over $81$ fixed shot-response contexts from three recovery-start slices. Shaded bands show standard error across fixed states or contexts. Effective support is $\exp(H)$ for the empirical categorical distribution, and distribution shift is Jensen--Shannon divergence from the initial checkpoint. The shot policy changes substantially relative to initialization while remaining locally sparse, whereas recovery expands its effective grid support over training and remains far from the initial distribution.}
    \label{fig:appendix-probe-coverage}
\end{figure*}

Figure~\ref{fig:appendix-probe-coverage} summarizes four aggregate statistics.
The top-left panel shows effective support for the joint shot-type--landing-zone distribution.
It stays close to one or two bins, indicating that the shot policy remains locally sparse under fixed contact states.
The top-right panel shows Jensen--Shannon divergence from the initial checkpoint for the same joint shot distribution.
The nonzero divergence indicates that self-play changes which sparse shot modes are preferred, rather than merely increasing random exploration.
At the latest controlled-contact checkpoint, the mean joint shot--landing Jensen--Shannon divergence is about $0.17$ nats, while the mean effective joint support remains near $1.3$ bins.

The bottom-left panel shows effective support over recovery-grid cells.
Unlike the shot support, recovery support grows more clearly, from about $1.69$ effective cells at initialization to about $2.15$ cells by $6.0$M steps.
The bottom-right panel shows that the recovery-grid distribution remains far from initialization, with mean Jensen--Shannon divergence reaching about $0.48$ nats by $6.0$M steps.
This pattern indicates that recovery does not collapse to a single default return-to-center target.
The learned policy spreads probability across multiple context-dependent recovery cells while maintaining a distribution that is consistently different from the initial policy.

The averages in Figure~\ref{fig:appendix-probe-coverage} are across fixed tactical states, not across naturally occurring rally states and not across independent training seeds.
For shots, each point averages the per-state metric over $27$ fixed contact states.
For recovery, each point averages the per-context metric over $81$ fixed shot-response contexts, obtained from three recovery-start slices.
The shaded bands are standard errors over these fixed states or contexts.
Thus the figure should be interpreted as an aggregate controlled-probe diagnostic: it shows how the learned conditional distributions change when the tactical inputs are held fixed.

\fi
\bibliography{references}

\end{document}